\documentclass[11pt,a4paper]{article}
\usepackage{authblk}
\usepackage[hyperref]{ranlp2023}
\aclfinalcopy

\usepackage{times}
\usepackage{latexsym}

\usepackage{txfonts}

\usepackage{amsmath}
\ifdefined\openbox
  \let\openbox\relax 
\fi
\usepackage{amsthm} % Now amsthm can define \openbox without error
\usepackage{amsthm}
\usepackage{float}
\usepackage{stfloats}
\usepackage{graphicx}
\usepackage{booktabs}
\usepackage[toc,page]{appendix}
\usepackage{microtype}

\title{ExPe: Exact Positional Encodings for Generative Transformer Models with Extrapolating Capabilities}

\author[1,2]{Aleksis Datseris}
\author[1]{Sylvia Vassileva}
\author[1]{Ivan Koychev}
\author[1,2]{Svetla Boytcheva}
\affil[1]{Faculty of Mathematics and Informatics, Sofia University St. Kliment Ohridski, Sofia, Bulgaria}
\affil[2]{Graphwise, Sofia, Bulgaria}
\affil[ ]{\textit{datseris@uni-sofia.bg}, \textit{\{svasileva, koychev\}@fmi.uni-sofia.bg}, \textit{svetla.boytcheva@graphwise.ai}}

\date{}

\begin{document}
\maketitle
\begin{abstract}
This paper introduces a novel approach to position embeddings in transformer models, named "Exact Positional Embeddings" (ExPE). An absolute positional embedding method that can extrapolate to sequences of lengths longer than the ones it was trained on. Traditional transformer models rely on absolute or relative position embeddings to incorporate positional information into token embeddings, which often struggle with extrapolation to sequences longer than those seen during training. Our proposed method utilizes a novel embedding strategy that encodes exact positional information by overriding specific dimensions of the embedding vectors, thereby enabling a more precise representation of token positions. The proposed approach not only maintains the integrity of the original embeddings but also enhances the model's ability to generalize to more extended sequences. In causal language modeling, our ExPE embeddings significantly reduce perplexity compared to rotary and sinusoidal embeddings, when tested on sequences longer than those used in training.
\end{abstract}

\section{Introduction}

The transformer architectures have revolutionized the field of natural language processing (NLP), enabling significant advancements in tasks such as machine translation, summarization, and question answering. Central to the success of transformers is the self-attention mechanism, which allows models to capture complex dependencies between tokens in a sequence. However, a critical challenge in leveraging self-attention is the incorporation of positional information, as the mechanism itself is inherently permutation invariant. 

Traditional approaches to position embeddings can be broadly categorized into two main types: absolute and relative methods. Absolute position embeddings, such as those introduced by \cite{Vaswani:2017}, typically involve adding vectors to the token embeddings. The vector could be fixed or learned, allowing the model to understand the position of each token in the sequence. Although effective, these methods often struggle with generalization to sequences longer than those encountered during training. On the other hand, relative position embeddings, as explored by \cite{Shaw2018SelfAttentionWR} and others, focus on encoding the relative distances between tokens, offering improved flexibility but still facing challenges in capturing precise positional information.

The main issues with the inability to generalize to sequences longer than those seen during training are multi-fold. First, the computational requirements to train a neural network are an order of magnitude larger than the computational resources required for inference with the same neural network. Second, there is an ever-growing need for longer context models. Research has led to the development of models with context lengths of up to $2$ million tokens \cite{ding2024longropeextendingllmcontext}, and that is still not enough for the commercial and research needs we have for those models, since there are many tasks that require more tokens to be solved. Third, and most importantly, is the fact that the primary means by which those models are so effective is the self-attention mechanism, which has quadratic complexity in both time and space. This makes it economically and computationally inefficient for us to scale the context length of those models. One would think that there isn't a fundamental difference in how language is processed and understood as the sequences become longer, or at least the understanding of where the words are in the sequence shouldn't be an issue. What we explore in this paper is whether there are ways to give the models positional information in a way that they can generalize to sequences longer than the ones seen during training. If achieved, it could reduce the cost, time, and environmental impact of implementing such models by an order of magnitude.

\section{Background and Proposed Approach}
\label{sec:approach}

    Extensive background and related work can be found in \ref{background}. The attention mechanism \cite{Vaswani:2017} defined in \ref{fn:qkv0} is permutation invariant, meaning that it has no positional information about where the tokens are in the sequence. 

\begin{equation}
\begin{aligned}
		\mathbf{q}_m &=f_q(\mathbf{x}_m) = W_qx_m\\
		\mathbf{k}_n &=f_k(\mathbf{x}_n) = W_kx_n\\
		\mathbf{v}_n &=f_v(\mathbf{x}_n) = W_vx_n\\
           & W_{t:t\in\{q,k,v\}}  \in  \mathbb{R}^{d\times d} \\
            \mathrm{Attention}(q_m, k_n, v_n) &= \mathrm{softmax}(\frac{\mathbf{q}_m\mathbf{k}_n^T}{\sqrt{d}})\mathbf{v}_n \\
	\end{aligned}
	\label{fn:qkv0}
\end{equation}
    
    Traditional methods for positional encodings either add/apply the information to each embedding or add/apply it to the query ($\mathbf{q}_n =f_q(\mathbf{x}_n)$) and key ($\mathbf{k}_n =f_k(\mathbf{x}_n)$) embeddings. 

\subsection{Formulation of ExPE}

\begin{figure*}[hb!]
	\centering
	\includegraphics[width=0.75\textwidth]{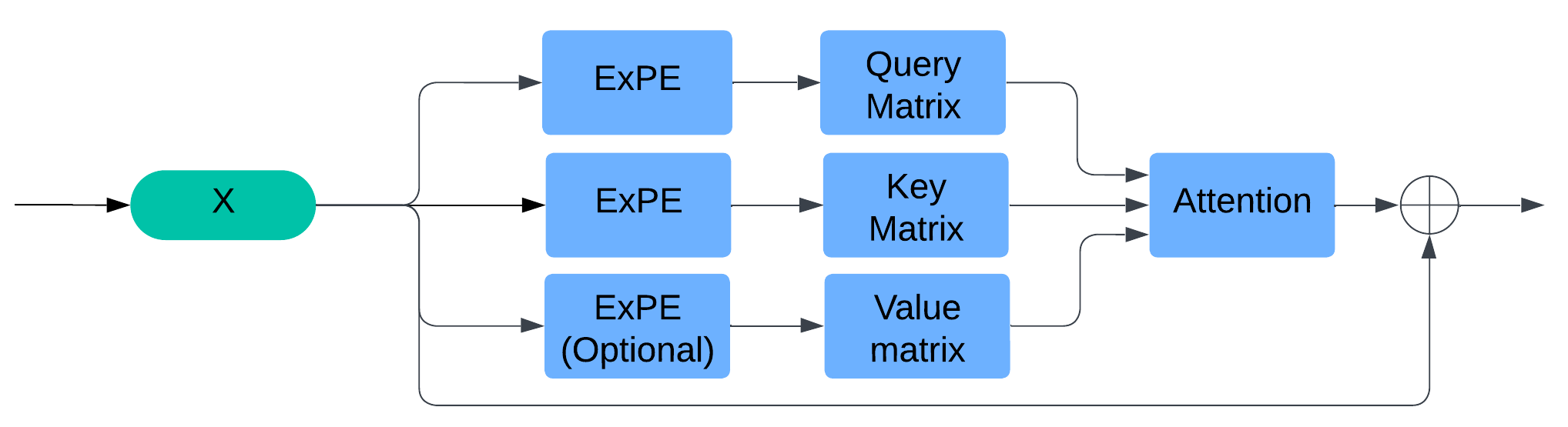}
	\caption{Flowchart showing when ExPE is applied in self-attention.}
	\label{fig:X_pos_flow}
\end{figure*}

    To address the issue with the inability of current approaches to extrapolate to sequences longer than the ones seen during training, we present Exact Positional Encodings (ExPE). In contrast, other approaches attempt to incorporate positional information by augmenting the embedding vector with positional information or manipulating the vector using techniques such as rotations. We propose the idea to override a limited number of values from the vector embedding with a vector representation of the position of the embedding in the sequence.

The main idea we propose is to override $l$ of the embedding dimensions to represent the exact position of the tokens. Starting with an initial value $S$ and increasing by a small constant $\theta$.

\begin{equation}
	\begin{aligned}
		\mathbf{q}_m &= f_q(\mathbf{x}_q, m) &= f_q(\varphi(x_q, m)) &= \mathbf{W}_{q}(\varphi(x_q, m)) \\
		\mathbf{k}_n &= f_k(\mathbf{x}_k, n) &= f_q(\varphi(x_k, n)) &= \mathbf{W}_{k}(\varphi(x_k, n))
	\end{aligned}
	\label{fn:deri-qk1}
\end{equation}

where $\mathbf{W}_{q}$ and $\mathbf{W}_{k}$ are the Query and Key matrices of the attention mechanism (\ref{fn:qkv}). The positional encoding ExPE written as $\varphi(x, i)$ is defined as:

\begin{equation}
	\begin{aligned}
            x &:=  (x_1, x_2, ..., x_d) \\
		\varphi(x, n) &:= (p_n, p_{n+1}, ..., p_{n+l-1}, x_{l+1}, ..., x_d),  \\
            p_n &:= S + \theta * n 
	\end{aligned}
	\label{fn:ExPE_def}
\end{equation}

Here $S$ is a defined constant, for example $0$, also $\theta$ is a predefined constant, for example $1/2m$, $d$ is the dimensionality of the model, and $m$ is the max expected length of input during training. The size of the positional encoding $l$ is also a hyperparameter whose size can vary.
The reason why we are overriding the first $l$ values in the vector instead of concatenating them is that the embeddings are applied before each transformer block, and that way, the size of the embeddings remains the same.
We have to note the fact that we override the first $l$ values of the vector, and consider them as a placeholder for the positional information.  
If token $x_n$ is at position $n$ and token $x_m$ is at position m after applying $\varphi()$ the first $l$ values in the $x_n$ will be $p_n, ..., p_{n+l-1}$ and $x_m$ will 
be $p_m, ..., p_{m+l-1}$ . The difference between $ p_n - p_m =  \theta * (n - m) $, so the 
greater the distance between two tokens, the greater the difference in the positional embedding values will increase proportionally. 

Since the fundamental principle behind this is that larger values should be interpreted as that the token is further from the start into the input sequence, not only exact and relative information about the positions of the token can be represented in this way, but it should allow the model to extrapolate for sequences that are longer than the inputs the model was trained on something other positional encodings fail to achieve.  A visualization to show better when exactly ExPE is applied to the query and keys is shown in Figure \ref{fig:X_pos_flow}. ExPE gets applied to the input of the Query Matrix and Key Matrix since those are the only embeddings that need positional information between them as the matrix multiplication between them will represent the attention between each pair of tokens in the sequence  $QK^T \in \mathbb{R}^{n \times n}$ the application to the input of the Value matrix is optional. The Query and Key embedding with encoded positional information, with the Value embeddings, go through the Scaled-Dot-Product attention operation, followed by the residual connection. The application of ExPE before the Query, Key, and Value matrices relies on the ability of those matrices to learn how to use ExPE's representation of positional information. Thanks to the residual connection, the information that ExPE overrides is not entirely lost, as it still travels through it.

\subsection{Rationale  Behind ExPE}
\label{Rationale  Behind ExPE}

    Figure \ref{fig:X_pos_example} in Appendix \ref{example appendix} shows a flowchart showing when ExPE is applied in self-attention.

    The application of ExPE before the Query, Key, and Value matrices in the Multi-Head Attention relies on the ability of the Multi-Head Attention to learn how to use ExPE's representation of positional information. Thanks to the residual connection, the information that ExPE overrides is not fully lost as it still travels through it.

    An intuitive way to think about it is to believe that the model will try to learn to hold the context-independent information in the first part of the vector and the context-dependent information in the second part, which ExPE does not override. For example if think about the word \textsl{"dog"} a dog could mean a German Shepherd an American Pit Bull Terrier or even a toy. Even though there is a big difference between the breeds, which are living, breathing creatures, and the toy, there is some context-independent information that unites all of those concepts, which the word \textsl{"dog"} could represent in our mind. This context-independent information does not need to be contextualized, so if it gets overridden by ExPE, it will not be lost, as shown before, and it can still change thanks to the contextualized information coming from the attention mechanism. 

\subsection{ExQPE Quantization Stable Alternative}

    In fp32 there are about $1e9$ numbers between $[0, 1]$, in fp16 it's $33e6$, in TF32  this drops to about $1e6$ and in bf16 there are only around $16e3$. Which could be problematic when needing a longer context and using heavy quantization. To deal with this, we developed a second version of ExPE called ExQPE, Exact quantizable positional encodings. The idea behind ExQPE is to increment only one of the $l$ dimensions of the positional encodings.
    \begin{equation}
	\begin{aligned}
		\mathbf{q}_m &= f_q(\mathbf{x}_q, m) &= f_q(\varphi(x_q, m)) &= \mathbf{W}_{q}(\varphi(x_q, m)) \\
		\mathbf{k}_n &= f_k(\mathbf{x}_k, n) &= f_q(\varphi(x_k, n)) &= \mathbf{W}_{k}(\varphi(x_k, n))
	\end{aligned}
	\label{fn:deri-qk2}
\end{equation}

where  $S, \theta_1, \theta_2$ are hyperparameters and
\begin{equation}
	\begin{aligned}
            &x :=  (x_1, x_2, ..., x_d) \\
		&\varphi(x, n) := (p_{n,0}, p_{n,1}, ..., p_{n,l-1}, x_{l}, ..., x_d),  \\
            &p_{0} := (S+0*\theta_1 + \theta_2, \\
            &S+1*\theta_1, ...,S+(l-1)*\theta1) \\
            &p_{0, i, i\neq0} := S+i*\theta1 \\
            &p_{k, i} := p_{k}[i] \\
            &p_{i} := (p_{i-1, 0}, p_{i-1, 1}, ..., p_{i-1, k-1}, p_{i-1, k} + \theta_2,\\
            & p_{i-1, k+1}, ...,p_{i-1, l-1})\ (where\ k \equiv i \mod l )\\
	\end{aligned}
	\label{fn:deri-XQ}
\end{equation}

  %  Here $S, \theta_1, \theta_2$ are hyperparameters.

\section{Data}
\label{Data}

    For the proper training of a language model, high-quality text data is necessary. There can't be specified an exact amount of data, but based on \cite{hoffmann2022trainingcomputeoptimallargelanguage}, at least $20$ tokens per parameter offer a good compute-to-performance ratio. 

\subsection{Data for Causal Language Modelling and Masked Language Modelling}
\label{Small data}
    The data used for the initial experiments on causal language modeling and masked language is a subset taken from the Fineweb dataset \cite{penedo2024finewebdatasetsdecantingweb}. The FineWeb dataset comprises over 15 trillion tokens of cleaned and deduplicated English web data from CommonCrawl. From that dataset, a subset was selected that contained only text at least 4000 tokens long, until a total of 700 million tokens were accumulated. The texts are very complex and vary widely in topics. The llama \cite{touvron2023llamaopenefficientfoundation} tokenizer was used for tokenization. The total data collected was $700$M tokens, of which $100$M was equally divided into the dev/test split. The average length of the individual texts was $\ 8000$ tokens. For more details, look at Appendix \ref{small data appendix}.
    
\subsection{Data for Bigger Causal Language Models}
\label{Large data}
For developing a bigger GPT model, we decided to use a more standard pertaining scheme. We used the Fineweb-edu dataset \cite{lozhkov2024fineweb-edu}, a subset of 1.3T tokens from educational web pages filtered from the FineWeb dataset. A random subset of $10$ billion parameters was selected from the dataset. The average length of the texts was $1000$ tokens and the total tokens used for training was $9.8$B the other were equally divided into the dev/test set (Appendix \ref{Large data appendix}).

    For these models, we will also evaluate them on standard LLM benchmarks for assessing language understanding and reasoning.

\section{Experiments and Results}

\subsection{Experiments and Evaluation of Causal Language Modeling}
\label{small scale experiments}
    For comparison of out method we will focus mostly on RoPE \cite{su2023roformerenhancedtransformerrotary} as they are the state-of-the-art technique for positional encodings, but for historic reasons, we will do some evaluation on the Sinosoidal positional encodings \cite{Vaswani:2017} to reevaluate that RoPE is a superior technique.

\textbf{Experimental Setup}. The data used is described in section \ref{Small data}. The hardware used for training was a single A40 48GB GPU. The models were trained on sequences of 512 tokens due to resource limitations. The list of training parameters can be found in Table \ref{Trainig setup} in Appendix \ref{small data appendix}.

    Both models had $35$M parameters. The parameters were based on the parameters of the GPT models from \cite{brown2020languagemodelsfewshotlearners}. Since the model is 4 times smaller than the smallest GPT model from the paper, we decided to reduce the number of layers by half and the embedding dimension by half to achieve the desired size. We did not use hyperparameter searches due to computational resource limitations. The model parameters can be found in Table \ref{GPT parameters table} in Appendix \ref{small data appendix}.
    Two models were compared: one standard decoder transformer model with Sinusoidal PE, one with rotary embeddings, and one that uses the proposed positional embeddings called ExPE for short.

\textbf{Results}. The experiments with this approach yielded a good result (Table \ref{Exact positional small results}) compared to RoPE, and the model's cross-entropy also remained stable when given inputs that were $2$ or $4$ times longer than the training data text, which was $512$ tokens in length. This is something that relative, rotary, and sinusoidal positional encoding don't achieve. Our model showed even a slight improvement when the input length was increased. Our explanation for that is that the texts are initially at least $4000$ tokens, and the majority of input sequences that we test the model with are cut from a longer text. There is a lot of missing context, as our model can extrapolate positions for more extended sequences. We not only don't see the perplexity increase, but it even decreases slightly.

\label{Ablation studies}

\begin{table*}[t]
\centering
\begin{tabular}{lcccc}
\toprule
\textbf{Model}           & \textbf{Loss (ev=1)} & \textbf{Loss (ev=2)}  & \textbf{Loss (ev=4)} & \textbf{Training time} 
\\ \midrule
LLama Small          & 2.89             &  3.80     &  4.95        & 1.00                        \\
ExPE Small               & 2.87              & \textbf{2.83}   & \textbf{3.25}  &  \textbf{0.93}                          \\
ExQPE Small               & \textbf{2.86}              & \textbf{2.83}   & 3.35  &  \textbf{0.93}                          \\
\midrule
LLama Medium          & \textbf{2.63}             &  3.45     &  4.55   & 1.00                                  \\
ExPE Medium               & \textbf{2.63}              & \textbf{2.59}   & 2.71 & \textbf{0.78}                       \\
ExQPE Medium & \textbf{2.63}             & \textbf{2.59}   & \textbf{2.68}  & \textbf{0.78} \\

\bottomrule
\end{tabular}
\caption{
The legend (ev=1) is the evaluation on texts of length the training length; here, for all models, the training length is 512 tokens. (ev=2) is the evaluation on texts with length double the training length, etc.
}
\label{Large Models results}
\end{table*}

    \textbf{Ablation Studies on ExPE}. We did a series of ablation studies to verify that all aspects of the ExPE positional encodings are necessary. The results can be seen in Table \ref{ab studies} in Appendix \ref{small data appendix}. 
    
    \textbf{Do we need to increase the values we override in the same vector instead of overriding with the same value?} Increasing the value in the vector we override seems to give a significant performance improvement. This could also be to a large extent because initializing two values in embedding with the same value makes them learn the same thing due to the way the gradient flows through \cite{kumar2017weightinitializationdeepneural}. 
    
    \textbf{Do you need to override more than one value, i.e., to have $l$ be greater than $1$.} With $l=1$ we achieve significantly worse results. The most significant factor should be simply the fact that the positional information is difficult to distinguish from the other embedding details when it is in a single dimension. 

    \textbf{Do we need to apply ExPE on every transformer block?} Similarly to how RoPe needs to be used in each transformer block, the same applies to ExPE. It appears that the model struggles to retain the positional information in the embeddings after they are passed through the layers. Therefore, for the model to utilize the position information, it must be applied in each transformer block. 

    \textbf{Do we need to fix $S$ and $\theta$ in ExPE or they can be learned?}
    Since the model can use the fixed values, one would think that the model could learn the values for $S$ and $\theta$. Here we see that the model fails to learn basic language modeling. It seems that having only two learned parameters affects so much of the output, making the model unstable and making it difficult for the model to learn.   

    In \cite{gu2024mambalineartimesequencemodeling} used a specific initialization was used to initialize their model in a stable state. We decide to make $S$ and $\theta$ learnable, but instead of initializing them randomly, initialize them with the same values as the ones in the non learned variant of the model.  While achieving better results compared to the previous method, the training time was over $60\%$, making the increased computational cost not worth the slight, arguable improvement.
\begin{table}[H]
\begin{center}
\begin{tabular}{@{}lccc@{}}
\toprule
\textbf{} & \textbf{Sinusoidal} & \textbf{RoPe} & \textbf{ExPE} \\ \midrule
\textbf{Loss (ev=1)} & 4.0 & \textbf{3.88} & 3.93 \\
\textbf{Loss (ev=2)} & 4.75 & 4.37 & \textbf{3.87} \\
\textbf{Loss (ev=4)} & 5.64 & 5.05 & \textbf{3.88} \\
\bottomrule
\end{tabular}
\end{center}
\caption{The legend (ev=1) is the evaluation on texts with length the training length; here, for all models, the training length is 512 tokens. (ev=2) is the evaluation on texts with length double the training length, etc. \newline
}
\label{Exact positional small results}
\end{table}

\subsection{Large-scale Experiments}
\label{Large-scale Experiments}
 The data used is described in section \ref{Large data}. We tokenized the dataset using Mistralai's Mistral-Small-Instruct-2409 \cite{jiang2023mistral7b} tokenizer. The architecture used was the Llama 3 architecture without the key-value cache, since it only increases inference speed. The embedding weights of the medium models are shared with their linear units.

 The model parameters found in Appendix \ref{Large data appendix} Table \ref{table:param} are based on the GPT parameters from \cite{brown2020languagemodelsfewshotlearners}.  The additional part of the training setup  can be found in Appendix \ref{Large data appendix}

    The results are shown in Table \ref{Large Models results}. What we see here is that ExPE still manages to maintain its length extrapolation capabilities while achieving comparable performance to RoPE, even at lengths seen during training. While also requiring significantly fewer computational resources.

\begin{table*}[t]
\begin{center}
\begin{tabular}{@{}lrrrrrr@{}}
\toprule
\textbf{Model}           & \textbf{Loss (ev=1)} & \textbf{Loss (ev=2)}  & \textbf{Loss (ev=4)} & \textbf{Loss (ev=8)}   & \textbf{Loss (ev=16)}
\\ \midrule
ExPE M             & \textbf{2.63}              & \textbf{2.59}   & 2.71 & 5.19 & 7.56       \\
ExPE M scaled by $0.5$          & 2.76             & 2.71   & \underline{2.68} & \underline{2.98} & \underline{5.65}      \\
\midrule
ExQPE M & \textbf{2.63}             & \textbf{2.59}   & 2.68 & 5.01 & 7.39        \\
ExQPE M scaled by $0.5$          & 2.74             & 2.69   & \textbf{2.65} & \textbf{2.76} & \textbf{4.98}      \\
\bottomrule
\end{tabular}
\end{center}
\caption{ All results are for ExPE Medium-sized models. The legend (ev=1) is the evaluation on texts of length equal to the training length. Here, for all models, the training length is 512 tokens. (ev=2) is the evaluation on texts with length double the training length, etc. 
}
\label{Scaling results}
\end{table*}

    We also evaluated the performance of the models on the HellaSwag \cite{zellers2019hellaswagmachinereallyfinish}, MMLU(Measuring Massive Multitask Language Understanding) \cite{hendrycks2021measuringmassivemultitasklanguage}, ARC and ARC easy \cite{moskvichev2023conceptarcbenchmarkevaluatingunderstanding} benchmarks to test how they perform on standard LLM benchmarks and we see that their results (found in Table \ref{Large Models results HellaSwag}) are comparable. In terms of training inference speed and memory requirements, ExPE is practically equivalent to Sinusoidal, as shown by \cite{press2022trainshorttestlong}; they are significantly faster than RoPE. In general, even if ExPE doesn't demonstrate a remarkable ability to extrapolate to sequences longer than those seen during training, it still shows significant improvement compared to RoPe and Sinusoidal. Also manages to maintain performance compared to RoPe while requiring significantly less compute when compared to sequence lengths seen during training.
\begin{table}[H]
\begin{center}
\begin{tabular}{@{}lrrrr@{}}
\toprule
\textbf{Model}        & \textbf{HS}  & \textbf{MMLU} & \textbf{ARC} & \textbf{A E}
\\ \midrule
LLama S          & \textbf{0.29}       & \textbf{0.26}      & \textbf{0.24}   & \textbf{0.30}                           \\
ExPE S               & 0.28    & \textbf{0.26}  & \textbf{0.24}      & 0.29                             \\
ExQPE S               & 0.27    & \textbf{0.26}  & \textbf{0.24}      & 0.29                             \\

\midrule
LLama M          & 0.31     & \textbf{0.26}    & \textbf{0.26}  & 0.30                                \\
ExPE M               & 0.31               & \textbf{0.26}    & 0.25  & 0.30                    \\
ExPE M ($0.5$)           & 0.31               & \textbf{0.26}    & \textbf{0.25}  & 0.30                    \\
ExQPE M               & \textbf{0.32}               & \textbf{0.26}    & 0.25  & \textbf{0.31}                    \\
ExQPE M ($0.5$)          & \textbf{0.32}            & \textbf{0.26}  & \textbf{0.25} & \textbf{0.31}  \\
\bottomrule
\end{tabular}
\end{center}
\caption{ Performance of the small (S) and medium (M) sized models on HellaSwag (HS), MMLU, ARC, ARC Easy (A E). The notation $<$Model ($x$)$>$ means that the model's encodings are scaled by a factor of $x$.
}
\label{Large Models results HellaSwag}
\end{table}
\subsection{Scaling the Encoding}
 \cite{chen2023extendingcontextwindowlarge} showed that if a model uses RoPe at a certain frequency and a certain context length, that model context length can be easily extended by scaling the frequency. If you would like to double the context length, you could scale the frequency of RoPe by half and do additional training. The method is effective because it requires a small amount of training for the model to extrapolate to the new training length. The issue is that you still need to do additional training. We decided to check how well ExPE behaves when scaling the encoding values without additional training. When we trained the model with a context length of $512$ tokens during training the first positions of ExPE values ranged from $0$ to $0.25$ while when texting to lengths of 2048 tokens those values ranged from  $0$ to $1$ so we decided to scale them by half to see how will the model behave without additional training.

    The results shown in tables \ref{Scaling results} and \ref{Large Models results HellaSwag} show that for the longest extrapolation, the scaling helps, but for lengths seen during training, it slightly decreases performance, and interestingly, for lengths twice the training length, we see the same. This suggests that scaling could be used to enhance performance for more extended periods without requiring additional training. For even longer extrapolations, further training with scaling on longer context lengths should also yield good results with minimal extra training steps.

\section{Limitations}
\label{limits}
    Due to technical limitations, the models trained in this work are too small and were trained on a too short context length, making it difficult for us to reliably state that ExPE and ExQPE are viable state-of-the-art techniques for positional encodings in transformers. Additionally, here we only compare the results of models which have gone through only the pretraining stage and have not gone through the instruction tuning phase with supervised fine-tuning, DPO \cite{rafailov2024directpreferenceoptimizationlanguage} or Reinforcement learning from human feedback(RLHF) \cite{ouyang2022traininglanguagemodelsfollow}. Finally, a general issue with longer context is that human language is inherently local due to the limitations of the human brain; a person can only follow a few sentences at a time. Therefore, many long documents still lack long dependencies \cite{yang2025qwen251mtechnicalreport}. The development of both proper benchmarks and training data tailored explicitly to long context dependencies is necessary for us to properly test and train the extended context capabilities of language models.

\section{Conclusion}

    In this work, we introduced Exact Position Embeddings (ExPE) and ExQPE, two novel positional encoding methods designed to improve the extrapolation capabilities of transformer models. By explicitly encoding position information precisely, ExPE and ExQPE enhance the model's capabilities for length extrapolation of sequences longer than those seen during training. Our experiments demonstrated that ExPE and ExQPE outperform traditional sinusoidal embeddings and achieve competitive results compared to Rotary Position Embeddings (RoPE), while being more efficient.

    We demonstrated that ExPE and ExQPE effectively retain position information and adapt to more extended sequences without requiring additional training for the causal language modeling task.
    We also demonstrated that, with fixed scaling of the two approaches, we can further enhance the length extrapolation capabilities of the model without additional training. Our results indicate that ExPE and ExQPE present a promising alternative to existing positional encoding techniques.

\textbf{Directions for Future Development} 
     For ExPE and ExQPE to be proven as practical techniques for positional encodings, larger experiments should be conducted with more extensive models, greater data, and longer context lengths. The development of proper benchmarks and training data tailored explicitly to long context dependencies is necessary for us to properly test the extended context capabilities of language models. Comparing the method to techniques like Dual Chunk
    Attention \cite{an2024trainingfreelongcontextscalinglarge} and Position Interpolation \cite{chen2023extendingcontextwindowlarge} should be done.

\newpage

\section*{Acknowledgments}
This work is partially supported by the project UNITe BG16RFPR002-1.014-0004 funded by PRIDST.

We are extremely grateful to Prof. DSc. Stoyan Mihov, Institute of Information and Communication Technologies, Bulgarian Academy of Sciences, for giving us initial guidance and helping us define the task at hand when developing this method.

\bibliographystyle{acl_natbib}
\bibliography{anthology,ranlp2023}

\begin{thebibliography}{29}
\expandafter\ifx\csname natexlab\endcsname\relax\def\natexlab#1{#1}\fi

\bibitem[{An et~al.(2024)An, Huang, Zhang, Gong et~al.}]{an2024trainingfreelongcontextscalinglarge}
Chenxin An, Fei Huang, Jun Zhang, Shansan Gong, et~al. 2024.
\newblock \href {http://arxiv.org/abs/2402.17463} {Training-free long-context scaling of large language models}.

\bibitem[{Brown et~al.(2020)Brown, Mann, Ryder, Subbiah et~al.}]{brown2020languagemodelsfewshotlearners}
Tom~B. Brown, Benjamin Mann, Nick Ryder, Melanie Subbiah, et~al. 2020.
\newblock \href {http://arxiv.org/abs/2005.14165} {Language models are few-shot learners}.

\bibitem[{Chen et~al.(2023)Chen, Wong, Chen, and Tian}]{chen2023extendingcontextwindowlarge}
Shouyuan Chen, Sherman Wong, Liangjian Chen, and Yuandong Tian. 2023.
\newblock \href {http://arxiv.org/abs/2306.15595} {Extending context window of large language models via positional interpolation}.

\bibitem[{DeepSeek-AI et~al.(2025)DeepSeek-AI, Liu, Feng, Xue et~al.}]{deepseekai2025deepseekv3technicalreport}
DeepSeek-AI, Aixin Liu, Bei Feng, Bing Xue, et~al. 2025.
\newblock \href {http://arxiv.org/abs/2412.19437} {Deepseek-v3 technical report}.

\bibitem[{Devlin et~al.(2019)Devlin, Chang, Lee, and Toutanova}]{Devlin2019BERTPO}
J.~Devlin, Ming-Wei Chang, Kenton Lee, and Kristina Toutanova. 2019.
\newblock Bert: Pre-training of deep bidirectional transformers for language understanding.
\newblock In \emph{NAACL-HLT}.

\bibitem[{Ding et~al.(2024)Ding, Zhang, Zhang, Xu et~al.}]{ding2024longropeextendingllmcontext}
Yiran Ding, Li~Lyna Zhang, Chengruidong Zhang, Yuanyuan Xu, et~al. 2024.
\newblock \href {http://arxiv.org/abs/2402.13753} {Longrope: Extending llm context window beyond 2 million tokens}.

\bibitem[{Dubey et~al.(2024)Dubey, Jauhri, Pandey, Kadian et~al.}]{dubey2024llama3herdmodels}
Abhimanyu Dubey, Abhinav Jauhri, Abhinav Pandey, Abhishek Kadian, et~al. 2024.
\newblock \href {http://arxiv.org/abs/2407.21783} {The llama 3 herd of models}.

\bibitem[{Gu and Dao(2024)}]{gu2024mambalineartimesequencemodeling}
Albert Gu and Tri Dao. 2024.
\newblock \href {http://arxiv.org/abs/2312.00752} {Mamba: Linear-time sequence modeling with selective state spaces}.

\bibitem[{Haviv et~al.(2022)Haviv, Ram, Press, Izsak, and Levy}]{haviv-etal-2022-transformer}
Adi Haviv, Ori Ram, Ofir Press, Peter Izsak, and Omer Levy. 2022.
\newblock \href {https://doi.org/10.18653/v1/2022.findings-emnlp.99} {Transformer language models without positional encodings still learn positional information}.
\newblock In \emph{Findings of the Association for Computational Linguistics: EMNLP 2022}, pages 1382--1390, Abu Dhabi, United Arab Emirates. Association for Computational Linguistics.

\bibitem[{Hendrycks et~al.(2021)Hendrycks, Burns, Basart, Zou et~al.}]{hendrycks2021measuringmassivemultitasklanguage}
Dan Hendrycks, Collin Burns, Steven Basart, Andy Zou, et~al. 2021.
\newblock \href {http://arxiv.org/abs/2009.03300} {Measuring massive multitask language understanding}.

\bibitem[{Hoffmann et~al.(2022)Hoffmann, Borgeaud, Mensch, Buchatskaya et~al.}]{hoffmann2022trainingcomputeoptimallargelanguage}
Jordan Hoffmann, Sebastian Borgeaud, Arthur Mensch, Elena Buchatskaya, et~al. 2022.
\newblock \href {http://arxiv.org/abs/2203.15556} {Training compute-optimal large language models}.

\bibitem[{Jiang et~al.(2023)Jiang, Sablayrolles, Mensch, Bamford et~al.}]{jiang2023mistral7b}
Albert~Q. Jiang, Alexandre Sablayrolles, Arthur Mensch, Chris Bamford, et~al. 2023.
\newblock \href {http://arxiv.org/abs/2310.06825} {Mistral 7b}.

\bibitem[{Kumar(2017)}]{kumar2017weightinitializationdeepneural}
Siddharth~Krishna Kumar. 2017.
\newblock \href {http://arxiv.org/abs/1704.08863} {On weight initialization in deep neural networks}.

\bibitem[{Lozhkov et~al.(2024)Lozhkov, Ben~Allal, von Werra, and Wolf}]{lozhkov2024fineweb-edu}
Anton Lozhkov, Loubna Ben~Allal, Leandro von Werra, and Thomas Wolf. 2024.
\newblock \href {https://doi.org/10.57967/hf/2497} {Fineweb-edu: the finest collection of educational content}.

\bibitem[{Moskvichev et~al.(2023)Moskvichev, Odouard, and Mitchell}]{moskvichev2023conceptarcbenchmarkevaluatingunderstanding}
Arseny Moskvichev, Victor~Vikram Odouard, and Melanie Mitchell. 2023.
\newblock \href {http://arxiv.org/abs/2305.07141} {The conceptarc benchmark: Evaluating understanding and generalization in the arc domain}.

\bibitem[{Ouyang et~al.(2022)Ouyang, Wu, Jiang, Almeida et~al.}]{ouyang2022traininglanguagemodelsfollow}
Long Ouyang, Jeff Wu, Xu~Jiang, Diogo Almeida, et~al. 2022.
\newblock \href {http://arxiv.org/abs/2203.02155} {Training language models to follow instructions with human feedback}.

\bibitem[{Penedo et~al.(2024)Penedo, Kydlíček, allal, Lozhkov et~al.}]{penedo2024finewebdatasetsdecantingweb}
Guilherme Penedo, Hynek Kydlíček, Loubna~Ben allal, Anton Lozhkov, et~al. 2024.
\newblock \href {http://arxiv.org/abs/2406.17557} {The fineweb datasets: Decanting the web for the finest text data at scale}.

\bibitem[{Press et~al.(2022)Press, Smith, and Lewis}]{press2022trainshorttestlong}
Ofir Press, Noah~A. Smith, and Mike Lewis. 2022.
\newblock \href {http://arxiv.org/abs/2108.12409} {Train short, test long: Attention with linear biases enables input length extrapolation}.

\bibitem[{Qwen et~al.(2025)Qwen, Yang, Yang, Zhang et~al.}]{qwen2025qwen25technicalreport}
Qwen, An~Yang, Baosong Yang, Beichen Zhang, et~al. 2025.
\newblock \href {http://arxiv.org/abs/2412.15115} {Qwen2.5 technical report}.

\bibitem[{Rafailov et~al.(2024)Rafailov, Sharma, Mitchell, Ermon et~al.}]{rafailov2024directpreferenceoptimizationlanguage}
Rafael Rafailov, Archit Sharma, Eric Mitchell, Stefano Ermon, et~al. 2024.
\newblock \href {http://arxiv.org/abs/2305.18290} {Direct preference optimization: Your language model is secretly a reward model}.

\bibitem[{Raffel et~al.(2023)Raffel, Shazeer, Roberts et~al.}]{raffel2023exploringlimitstransferlearning}
Colin Raffel, Noam Shazeer, Adam Roberts, et~al. 2023.
\newblock \href {http://arxiv.org/abs/1910.10683} {Exploring the limits of transfer learning with a unified text-to-text transformer}.

\bibitem[{Shaw et~al.(2018)Shaw, Uszkoreit, and Vaswani}]{Shaw2018SelfAttentionWR}
Peter Shaw, Jakob Uszkoreit, and Ashish Vaswani. 2018.
\newblock Self-attention with relative position representations.
\newblock In \emph{NAACL-HLT}.

\bibitem[{Su et~al.(2023)Su, Lu, Pan, Murtadha, Wen, and Liu}]{su2023roformerenhancedtransformerrotary}
Jianlin Su, Yu~Lu, Shengfeng Pan, Ahmed Murtadha, Bo~Wen, and Yunfeng Liu. 2023.
\newblock \href {http://arxiv.org/abs/2104.09864} {Roformer: Enhanced transformer with rotary position embedding}.

\bibitem[{Touvron et~al.(2023{\natexlab{a}})Touvron, Lavril, Izacard, Martinet et~al.}]{touvron2023llamaopenefficientfoundation}
Hugo Touvron, Thibaut Lavril, Gautier Izacard, Xavier Martinet, et~al. 2023{\natexlab{a}}.
\newblock \href {http://arxiv.org/abs/2302.13971} {Llama: Open and efficient foundation language models}.

\bibitem[{Touvron et~al.(2023{\natexlab{b}})Touvron, Martin, Stone, Albert et~al.}]{touvron2023llama2openfoundation}
Hugo Touvron, Louis Martin, Kevin Stone, Peter Albert, et~al. 2023{\natexlab{b}}.
\newblock \href {http://arxiv.org/abs/2307.09288} {Llama 2: Open foundation and fine-tuned chat models}.

\bibitem[{Vaswani et~al.(2017)Vaswani, Shazeer, Parmar, Uszkoreit, Jones, Gomez, Kaiser, and Polosukhin}]{Vaswani:2017}
Ashish Vaswani, Noam Shazeer, Niki Parmar, Jakob Uszkoreit, Llion Jones, Aidan~N Gomez, \L~ukasz Kaiser, and Illia Polosukhin. 2017.
\newblock \href {https://proceedings.neurips.cc/paper/2017/file/3f5ee243547dee91fbd053c1c4a845aa-Paper.pdf} {Attention is all you need}.
\newblock In \emph{Advances in Neural Information Processing Systems}, volume~30. Curran Associates, Inc.

\bibitem[{Yang et~al.(2025{\natexlab{a}})Yang, Li, Yang, Zhang et~al.}]{yang2025qwen3technicalreport}
An~Yang, Anfeng Li, Baosong Yang, Beichen Zhang, et~al. 2025{\natexlab{a}}.
\newblock \href {http://arxiv.org/abs/2505.09388} {Qwen3 technical report}.

\bibitem[{Yang et~al.(2025{\natexlab{b}})Yang, Yu, Li, Liu et~al.}]{yang2025qwen251mtechnicalreport}
An~Yang, Bowen Yu, Chengyuan Li, Dayiheng Liu, et~al. 2025{\natexlab{b}}.
\newblock \href {http://arxiv.org/abs/2501.15383} {Qwen2.5-1m technical report}.

\bibitem[{Zellers et~al.(2019)Zellers, Holtzman, Bisk, Farhadi, and Choi}]{zellers2019hellaswagmachinereallyfinish}
Rowan Zellers, Ari Holtzman, Yonatan Bisk, Ali Farhadi, and Yejin Choi. 2019.
\newblock \href {http://arxiv.org/abs/1905.07830} {Hellaswag: Can a machine really finish your sentence?}

\end{thebibliography}

\newpage
%\appendix

\begin{appendices}
\section{Appendix A}
\label{background}
\textbf{Preliminary}. The main mechanism that makes the transformer so effective is the Scaled Dot-Product Attention. Let's say that our model has a dimensionality of $d$, so each word embedding will be represented with a $ d$-dimensional vector. So, token embedding $w_i$ will be represented as a vector $x_i$, where $\mathbf{x}_i\in \mathbb{R}^{d}$. A sequence of $N$ tokens will be represented as $\mathbb{S}_N=\{w_i\}_{i=1}^{N}$ and their corresponding embeddings will be $x =\mathbb{E}_N = \{\mathbf{x}_i\}_{i=1}^{N} \ \mathbf{x}_i \in \mathbb{R}^{d}$. 
    
    The attention between two tokens $t_n$ and $t_m$ where $t_n$ is key and value and $t_m$ is the query will be 
    \begin{equation}
    \mathrm{Attention}(q_m, k_n, v_n) = \mathrm{softmax}(\frac{\mathbf{q}_m\mathbf{k}_n^T}{\sqrt{d}})\mathbf{v}_n
    \end{equation}
    Where 
    \begin{equation}
	\begin{aligned}
		\mathbf{q}_m &=f_q(\mathbf{x}_m) = W_qx_m\\
		\mathbf{k}_n &=f_k(\mathbf{x}_n) = W_kx_n\\
		\mathbf{v}_n &=f_v(\mathbf{x}_n) = W_vx_n\\
           & W_{t:t\in\{q,k,v\}}  \in  \mathbb{R}^{d\times d}
	\end{aligned}
	\label{fn:qkv}
\end{equation}

    The attention for the whole sequence will be

    \begin{equation}
   \mathrm{Attention}(Q, K, V) = \mathrm{softmax}(\frac{QK^T}{\sqrt{d_k}})V
   \label{fn:attention}
\end{equation}

Where

        \begin{equation}
	\begin{aligned}
		  Q &=f_q(\mathbf{x}) = W_q\mathbf{x}\\
		  K &=f_k(\mathbf{x}) = W_k\mathbf{x}\\
		  V &=f_v(\mathbf{x}) = W_v\mathbf{x}\\
           & W_{t:t\in\{q,k,v\}}  \in  \mathbb{R}^{d\times d} \\
   \end{aligned}
   \label{fn:attention2}
\end{equation}

    The fact that all the computations are to be done in parallel means that the mechanism has no positional information about where the tokens are in the sequence. So, for the encoder model, if we don't give the mechanism positional information. For decoder-only models \cite{haviv-etal-2022-transformer} showed that the causative nature of the attention mechanism can create an internal representation of the positional information, but adding positional information still significantly improves the model performance. So, for us to encode the positional information in the embeddings, we need to redefine $f_{t:t\in\{q,k,v\}}$ to take one more argument of the token position and to somehow encode it into the embedding.
    \begin{equation}
	\begin{aligned}
		\mathbf{q}_m &=f_q(\mathbf{x}_m, m)\\
		\mathbf{k}_n &=f_k(\mathbf{x}_n, n)\\
		\mathbf{v}_n &=f_v(\mathbf{x}_n, n)\\
	\end{aligned}
	\label{fn:qkv_pos}
\end{equation}
where $\mathbf{q}_m,\mathbf{k}_n$ and $ \mathbf{v}_n$ incorporate the $m^{th}$ and $n^{th}$ positions through $f_q,f_k$ and $f_v$, respectively. The query and key values are then used to compute the attention weights, while the output is computed as the weighted sum over the value representation. 
\begin{equation}
	\begin{aligned}
		a_{m,n}&=\frac{\exp(\frac{\mathbf{q}_m^{\intercal}\mathbf{k}_n}{\sqrt{d}})}{\sum_{j=1}^{N}\exp(\frac{\mathbf{q}_m^{\intercal}\mathbf{k}_j}{\sqrt{d}})}\\
		\mathbf{o}_m&=\sum_{n=1}^{N}a_{m,n}\mathbf{v}_{n}
	\end{aligned}
	\label{fn:attn}
\end{equation}

\textbf{Learned Absolute position embedding}
One of the first methods proposed by \cite{Vaswani:2017} is 
\begin{equation}
	f_{t:t\in\{q,k,v\}}(\mathbf{x}_i,i):=\mathbf{W}_{t:t\in\{q,k,v\}}(\mathbf{x}_i+e_i),
	\label{fn:adtv-posl}
\end{equation}
where $e_i\in\mathbb{R}^{d}$ is a d-dimensional embedding vector that is a learnable. The vector depends on the position of a token $\mathbf{x}_i$. The approach was proposed and tried by \cite{Vaswani:2017}. They demonstrated that it is capable of learning to represent positional information, but it has several disadvantages.
    
     One is that the model can't be used for inference on sequences longer than the ones it was trained on, since first, they don't exist, and second, even if we created new embeddings in the position embeddings, they still would need to be trained.
    Another significant disadvantage is that the new parameters increase the model's size and the computational requirements during training.
    
    Learned Absolute position embeddings were used by BERT \cite{Devlin2019BERTPO}. While back then $512$ tokens context length was a standard in recent years, a lot of models like llama 3\cite{dubey2024llama3herdmodels} have a context length of $128 000$ tokens, so for example for llama 3 $8$B \cite{dubey2024llama3herdmodels}. The use of such embeddings would add 0.5 billion training parameters. 
    
\textbf{Sinusoidal Absolute position embedding}. One would believe that there should be a method to give the model the information needed to infer the token's positions without the model needing to learn them. The approach \cite{Vaswani:2017} preferred is 
    \begin{equation}
	f_{t:t\in\{q,k,v\}}(\mathbf{x}_i,i):=\mathbf{W}_{t:t\in\{q,k,v\}}(\mathbf{x}_i+\mathbf{p}_i),
	\label{fn:adtv-posi}
    \end{equation}
    where $\mathbf{p}_i\in\mathbb{R}^{d}$ is a d-dimensional vector depending of the position of token $\mathbf{x}_i$ in the sequence. \cite{Vaswani:2017} have proposed to generate $\mathbf{p}_i$ using the sinusoidal function.
    \begin{equation}
	\begin{cases}
		\mathbf{p}_{i,2t}&=\sin(i/10000^{2t/d})\\
		\mathbf{p}_{i,2t+1}&=\cos(i/10000^{2t/d})
	\end{cases}
	\label{fn:sins}
    \end{equation}
in which $\mathbf{p}_{i,2t}$ is the $2t^{th}$ element of the d-dimensional vector $\mathbf{p}_i$.
    While this method didn't add any new parameters to the model, it still suffered from the model's inability to extrapolate to sequences longer than those seen during training.

\textbf{Relative position embeddings}. Both sinusoidal and learned position encodings focused on the absolute position of the token. Relative positional embeddings rely on a different approach, mainly the idea that the exact position of the word in the sequence is not relevant. What is essential is the word’s position relative to other words.

    \textbf{Rotary Position Embedding (RoPE)}. One of the most popular approaches is RoPE (Rotary Position Embeddings) proposed by \cite{su2023roformerenhancedtransformerrotary}. The approach is a relative positional embeddings method that again adds positional information to the query and key dot-product ($\mathbf{q}_m^{\intercal}\mathbf{k}_n$) values. To incorporate relative position information, it is required that the inner product of the query and key dot-product ($\mathbf{q}_m^{\intercal}\mathbf{k}_n$) values be formulated by a function $g$, 
    which takes only the word embeddings $\mathbf{x}_m$, $\mathbf{x}_n$, and their relative position $m -n$ as input variables. They propose that the query and key dot-product ($\mathbf{q}_m^{\intercal}\mathbf{k}_n$) values encode their position information only in the relative form:
\begin{equation}
	    \begin{aligned}
		    \langle\mathbf{q}_m,\mathbf{k}_n\rangle&=\langle f_q(\mathbf{x}_m, m),f_k(\mathbf{x}_n, n)\rangle\\
		    &=g(\mathbf{x}_m,\mathbf{x}_n,m-n)
		    \end{aligned}
	    \label{fn:formulation}
	\end{equation}

    They achieve this by applying a rotation matrix to the query and key tensors, which rotates the vectors relative to their position in the sequence.

\begin{equation}
\begin{aligned}
	\mathbf{q}_m^{\intercal}\mathbf{k}_n 
	&=(\mathbf{R}^d_{\Theta, m}\mathbf{W}_q\mathbf{x}_m)^\intercal(\mathbf{R}^d_{\Theta, n}\mathbf{W}_k\mathbf{x}_n) \\
    &=\mathbf{x}^\intercal\mathbf{W}_qR^d_{\Theta, n-m}\mathbf{W}_k\mathbf{x}_n
\end{aligned}    
\label{fn:rope-qk}
\end{equation}
where $\mathbf{R}^d_{\Theta, i}$ is the said $d$-dimensional rotation matrix and $\mathbf{R}^d_{\Theta, n-m}=(\mathbf{R}^{d}_{\Theta, m})^\intercal\mathbf{R}^d_{\Theta, n}$. The rotation matrix has two parameters $\Theta$, which is a pre-defined, non-zero constant (a common default value is $10,000$) that serves as a base for calculating the rotation angles applied to query and key vectors. The other parameter of the rotation matrix is the index of the token in the sequence. RoPE is currently the state-of-the-art positional embedding method for transformer-based models used in \cite{touvron2023llama2openfoundation}\cite{dubey2024llama3herdmodels}\cite{deepseekai2025deepseekv3technicalreport}\cite{qwen2025qwen25technicalreport}\cite{yang2025qwen3technicalreport}, although some of them use some techniques to help the extrapolation of the RoPE context length, like Dual Chunk
Attention \cite{an2024trainingfreelongcontextscalinglarge} or Position Interpolation \cite{chen2023extendingcontextwindowlarge}. Different methods have extrapolation abilities, but they haven't seen wide adoption like \cite{press2022trainshorttestlong} and \cite{raffel2023exploringlimitstransferlearning}.

\section{Appendix B}
\label{small data appendix}
    Training parameters for the small-scale models in section \ref{small scale experiments} are shown in Table \ref{Trainig setup}.

\begin{table}[H]
\begin{center}
\begin{tabular}{@{}lr@{}}
\toprule
\textbf{Parameter}              & \textbf{Values}                  \\ \midrule
\midrule
effective batch size            & $512$          \\
batch size                      & $64$ \\
gradient accumulation steps     & $8$ \\
peak learning rate              & $1e-3$   \\
end learning rate               & $3e-6$     \\
optimizer                       & AdamW                \\
optimizer $\beta_1$             & $0.9$                    \\
optimizer $\beta_2$             & $0.95$                    \\
weight decay                    & $0.1$                   \\
warm-up ratio                   & $10\%$                   \\
precision                       & fp$32$                  \\
\bottomrule
\end{tabular}
\end{center}
\caption{Training parameters for the small-scale models}
\label{Trainig setup}
\end{table}
\label{large training parameters}
    
\label{small scale models parameters}
\begin{table}[H]
\begin{center}
\begin{tabular}{@{}lr@{}}
\toprule
\textbf{Parameter}           & \textbf{Values}                  \\ \midrule
\midrule
vocab\_size          & 32 000          \\
seq\_len             & 512 \\
head\_size            & 32   \\
d\_model             & 384    \\
n\_heads             & 12                \\
n\_layers            & 6                     \\
dropout     & 0.1                   \\
total parameters    & 35 000 000                   \\
Non embedding parameters    & 23 000 000                   \\
\bottomrule
\end{tabular}
\end{center}
\caption{ Model parameters. The ExPE model uses positional encoding with a size of $l=24$. The standard decoders' positional encodings had a rope $\theta = 10,000$. Here, seq\_len means how many tokens long the sequences are that the model was trained on.}
\label{GPT parameters table}
\end{table}

    \begin{table*}[t]
    \begin{center}
    \begin{tabular}{@{}lrrrr@{}}
    \toprule
    \textbf{Model}           & \textbf{Loss (ev=1)} & \textbf{Loss (ev=2)} & \textbf{Loss (ev=4)}   &  \textbf{Epoch time}  \\ \midrule
    ExPE               & 3.93               & 3.87         & 4.26         & \textbf{1}                 \\
    ExPE (p is stable)                & 4.26               & 4.21        & \textbf{4.24}             & \textbf{1}         \\
    ExPE ($l=1$)                & 4.48               & 4.83        & 6.01                 & \textbf{1}     \\
    ExPE (learned initialized)                & \textbf{3.88}              & \textbf{3.82}       & 4.89     & 1.6              \\
    ExPE (learned)                & 7.33              & 7.34        & 7.61           & 1.6           \\
    ExPE (once)                & 4.43              & 4.82        & 6.18                  & \textbf{1}    \\
    
    \bottomrule
    \end{tabular}
    \end{center}
    \caption{Ablation studies done on ExPE. Here, for all models, the training length is 512 tokens. Where ExPE (p is stable) means we override with $p_i, p_i, .., p_i$ instead of  $p_i, p_{i+1}, .., p_{i+l}$. ExPE ($l=1$) uses $l=1$ for it's positional encodings. ExPE (once) has its positional encodings applied only once before the first transformer block.  ExPE (learned initialized) has its positional encodings $S$ and $\theta$ parameters learned but initialized with the same values as the non learned variant. ExPE (learned) has its positional encodings $S$ and $\theta$ parameters learned.
    }
    \label{ab studies}
    \end{table*}
\begin{figure*}[b]
	\centering
	\includegraphics[width=0.85\textwidth]{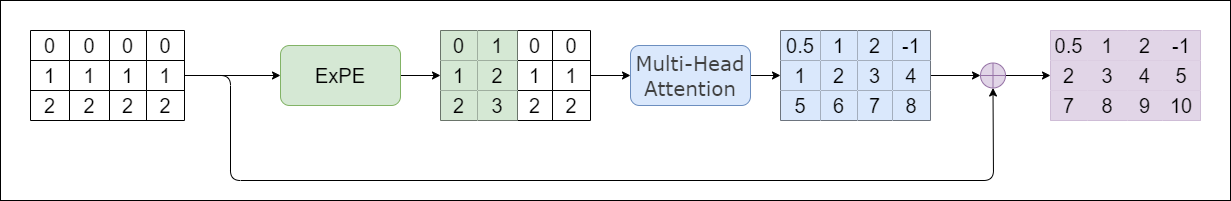}
	\caption{Flowchart showing when ExPE is applied in self-attention. ExPE gets applied to the input. Here, we assume self-attention, i.e., the Query, Key, and Value matrices are identical. }
	\label{fig:X_pos_example}
\end{figure*}

    The train, dev, and test data split for the experiments in section \ref{small scale experiments} are $600,000, 000$ training tokens and $50, 000, 000$ val/test, all with an average length of $8, 000$

\section{Appendix C}
\label{Large data appendix}
The list of model parameters mentioned in the large-scale experiments section \ref{Large-scale Experiments} can be found in table \ref{table:param}.
\begin{table}[H]
    \centering
    \begin{tabular}{lcc}
        \toprule
        & S Models & M Models \\
        \midrule
        $n_{\mathrm{params}}$    & 135M & 342M  \\
        $n_{\mathrm{layers}}$    & 12 & 24  \\
        $d_{\mathrm{model}}$     & 768 & 1024  \\
        $n_{\mathrm{heads}}$     & 12 & 16 \\
        $d_{\mathrm{head}}$      & 64 & 64  \\
        Peak learning rate       & $6.0 \times 10^{-4}$ & $3.0 \times 10^{-4}$  \\
        Rms norm                        & $1e-5$     & $1e-5$                      \\
        \bottomrule
    \end{tabular}
    \caption{Model parameters for the large-scale experiments models.}
    \label{table:param}
\end{table}

    The training data split for the large-scale experiments are $9 800 000 000$ training tokens and $100 000 000$ val/test, all with an average length of $1 000$

Training parameters for the large-scale experiments in section \ref{Large-scale Experiments} are shown in Table \ref{large Trainig setup}.

\begin{table}[H]
\begin{center}
\begin{tabular}{@{}lr@{}}
\toprule
\textbf{Parameter}              & \textbf{Values}                  \\ \midrule
\midrule
Training tokens  & 10B \\    
Batch Size               & 0.5M  \\
End lr               & $10\%$  of starting lr   \\
Lr scheduler         & cosine                \\
Optimizer                       & AdamW                \\
Optimizer $\beta_1$             & $0.9$                    \\
Optimizer $\beta_2$             & $0.95$                    \\
Weight decay                    & $0.1$                   \\
Warm-up ratio                   & $5\%$                   \\
Precision                       & bf$16$                  \\

$l$ (Only in ExPE \& ExQpe )             & $1/8$ of $d_{\mathrm{model}}$     \\
$S$ (Only in ExPE \& ExQpe )   & $0$     \\
$\theta$ (Only in ExPE )   & $1/2048$\\
$\theta_1$ (Only in ExQPE )   & $1/2048$\\
$\theta_2$ (Only in ExQPE )   & $1/16$\\
RoPe $\theta$ (Only in Llama)          & $10000$     \\
\bottomrule
\end{tabular}
\end{center}
\caption{Training and positional encodings parameters for the large-scale experiments models}
\label{large Trainig setup}
\end{table}

A table with a summary of the ablation studies done on ExPE is shown in Figure \ref{ab studies}.

\section{Appendix D}
\label{example appendix}
An Example of the application of ExPE is shown in Figure \ref{fig:X_pos_example}.

\end{appendices}

\end{document}